\documentclass[10pt,twocolumn,letterpaper]{article}

\usepackage{cvpr}              
\usepackage{graphicx}
\usepackage{amssymb}
\usepackage{pifont}
\usepackage{booktabs}
\usepackage{multirow}
\usepackage[autolanguage]{numprint}
\usepackage[normalem]{ulem}
\useunder{\uline}{\ul}{}
\usepackage{dirtytalk}
\usepackage{caption}
\usepackage[dvipsnames]{xcolor}

\definecolor{cvprblue}{rgb}{0.21,0.49,0.74}
\usepackage[pagebackref,breaklinks,colorlinks,citecolor=cvprblue]{hyperref}

\def\paperID{16717} 
\def\confName{CVPR}
\def\confYear{2024}

\title{$\mathbf{BoQ}$: A Place is Worth a Bag of Learnable Queries}

\author{Amar Ali-bey\thanks{Corresponding author} \qquad\quad Brahim Chaib-draa \qquad\quad  Philippe Gigu\`ere \\
{\normalsize Department of Computer Science and Software Engineering}\\
{\normalsize Université Laval, Québec, Canada}\\
{\tt\small amar.ali-bey.1@ulaval.ca, \{brahim.chaib-draa, philippe.giguere\}@ift.ulaval.ca}
}
\begin{document}
\maketitle
\begin{abstract}
In visual place recognition, accurately identifying and matching images of locations under varying environmental conditions and viewpoints remains a significant challenge. In this paper, we introduce a new technique, called Bag-of-Queries (BoQ), which learns a set of global queries, designed to capture universal place-specific attributes. Unlike existing techniques that employ self-attention and generate the queries directly from the input, BoQ employ distinct learnable global queries, which probe the input features via cross-attention, ensuring consistent information aggregation. In addition, this technique provides an interpretable attention mechanism and integrates with both CNN and Vision Transformer backbones. The performance of BoQ is demonstrated through extensive experiments on $14$ large-scale benchmarks. It consistently outperforms current state-of-the-art techniques including NetVLAD, MixVPR and EigenPlaces. Moreover, despite being a global retrieval technique (one-stage), BoQ surpasses two-stage retrieval methods, such as Patch-NetVLAD, TransVPR and R2Former, all while being orders of magnitude faster and more efficient. The code and model weights are publicly available at \url{https://github.com/amaralibey/Bag-of-Queries}.
\end{abstract}    
\section{Introduction}
\label{sec:intro}

\begin{figure}[t]
  \centering
   \includegraphics[width=1\linewidth]{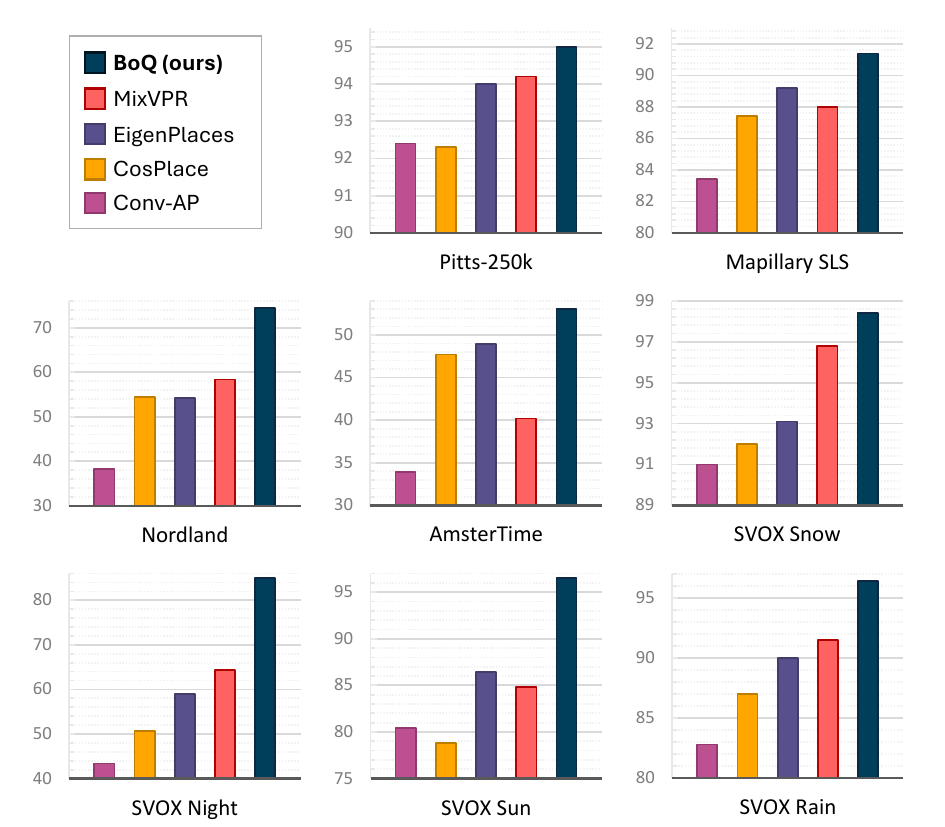}
   \caption{Recall@1 performance comparison between our proposed technique, Bag-of-Queries (BoQ), and current state of the art methods, Conv-AP~\cite{ali2022gsv}, CosPlace~\cite{berton2022rethinking}, MixVPR~\cite{ali2023mixvpr} and EigenPlaces~\cite{berton2023eigenplaces}. ResNet-50 is used as backbone for all techniques. BoQ consistently achieves better performance in various environment conditions such as viewpoint changes (Pitts-250k~\cite{torii2013visual}, MapillarySLS~\cite{warburg2020mapillary}), seasonal changes (Nordland~\cite{zaffar2021vpr}), historical locations (AmsterTime~\cite{yildiz2022amstertime}) and extreme lightning and weather conditions (SVOX~\cite{Berton_2021_svox}).}
   \label{fig:BoQ_block}
\end{figure}


Visual Place Recognition (VPR) consists of determining the geographical location of a place depicted in a given image, by comparing its visual features to a database of  previously visited places.
The dynamic and ever-changing nature of real-world environments pose significant challenges for VPR~\cite{masone2021survey, zhang2021visual}. Factors such as varying lighting conditions, seasonal changes and the presence of dynamic elements such as vehicles and pedestrians introduce considerable variability into the appearance of a place. Additionally, changes in viewpoint and image scale can expose previously obscured areas, further complicating the recognition process. These challenges are exacerbated by the operational constraints of VPR systems, which often need to operate in real-time and under limited memory. Consequently, there is a compelling need for efficient algorithms capable of generating compact yet robust representations.

With the rise of deep learning, numerous VPR-specific neural networks have been proposed~\cite{arandjelovic2016netvlad, kim2017learned, seymour2019semantically, liu2019stochastic, ge2020self, ali2022gsv, Ali-Bey_2022_BMVC, berton2022rethinking, ali2023mixvpr}, leveraging Convolutional Neural Networks (CNNs) to extract high-level features, followed by aggregation layers that consolidate these features into a single global descriptor. Such end-to-end trainable architectures have been instrumental in enhancing the efficiency and performance of VPR systems.

Recently, Vision Transformers (ViT)~\cite{dosovitskiy2020image} have demonstrated remarkable performance in various computer vision tasks, including image classification~\cite{chen2021crossvit}, object detection~\cite{carion2020end, liu2021swin} and semantic segmentation~\cite{zheng2021rethinking}. Their success can be attributed to their self-attention mechanism, which captures global dependencies between distant parts of the image~\cite{ghiasi2022vision}. However, current
Transformer-based VPR techniques~\cite{wang2022transvpr, zhu2023r2former, Zhang_2023_WACV, wang2023transformer}, often rely on \emph{reranking}~\cite{barbarani2023local}, a post-processing step aimed at refining the initial set of candidate locations identified through global descriptor search. The reranking process is usually done with geometric verification (\eg RANSAC) on stored local patch tokens, which is computationally and memory intensive. Moreover, the global retrieval process in Transformer-based methods, whether through specific aggregation of local patches~\cite{Zhang_2023_WACV, wang2022transvpr} or using the class token of the ViT~\cite{zhu2023r2former}, has yet to reach the performance levels of non-Transformer-based approaches like MixVPR~\cite{ali2023mixvpr} and CosPlace~\cite{zheng2021rethinking}.

In this paper, we bridge this performance gap, by introducing a novel Transformer-based aggregation technique, called Bag-of-Queries (BoQ), that learns a set of embeddings (global queries) and employs a cross-attention mechanism to probe local features coming from the backbone network. This approach enables each global query to consistently seek relevant information uniformly across input images. This is in contrast with  self-attention-based techniques~\cite{wang2022transvpr, Zhang_2023_WACV}, where the queries are dynamically derived from the input itself.
Furthermore, BoQ is designed for end-to-end training, thus seamlessly integrating with both conventional CNN and ViT backbones. Its effectiveness is validated through extensive experiments on multiple large-scale benchmarks, consistently outperforming state-of-the-art techniques, including MixVPR~\cite{ali2023mixvpr}, EigenPlaces~\cite{berton2023eigenplaces}, and NetVLAD~\cite{arandjelovic2016netvlad}. More importantly, BoQ, a single-stage (global) retrieval method that does not employ reranking, outperforms two-stage retrieval methods like TransVPR~\cite{wang2022transvpr} and R2Former~\cite{zhu2023r2former}, while being orders of magnitude more efficient in terms of computational and memory resources.
\section{Related Works}
\label{sec:related}
\begin{figure*}
  \centering
  \includegraphics[width=1\linewidth]{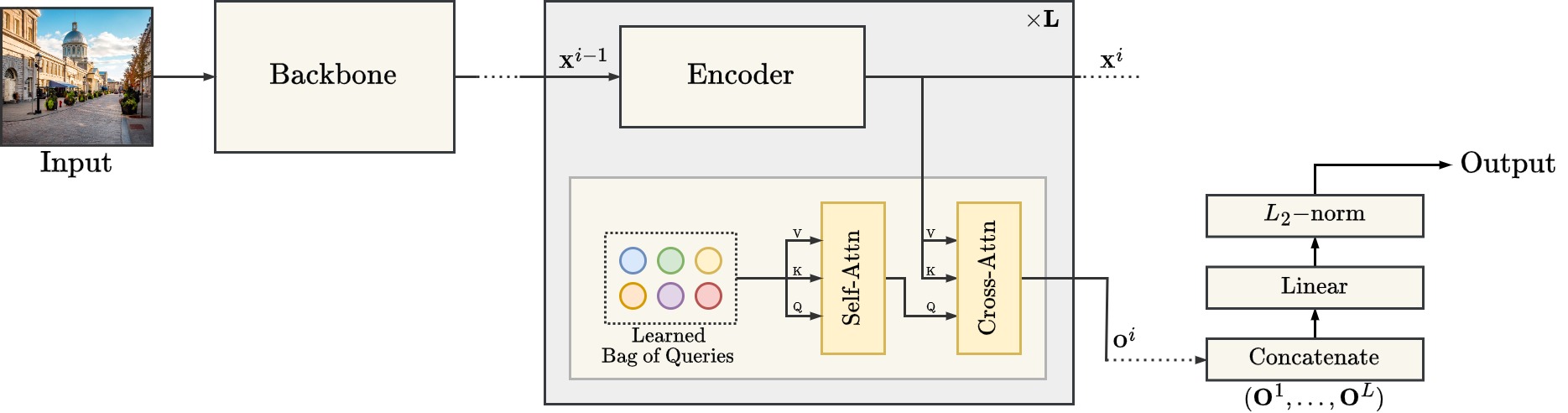}
  \caption{Overall architecture of the Bag-of-Queries (BoQ) model. The input image is first processed by a backbone network to extract its local features, which are then sequentially refined in a cascade of Encoder units. Each BoQ block contains a set of learnable queries $\mathbf{Q}$ (Learned Bag of Queries), which undergo self-attention to integrate their shared information. The refined features $\mathbf{X}^i$ are then processed through cross-attention with $\mathbf{Q}$ for selective aggregation. Outputs from all BoQ blocks $(\mathbf{O}^1, \mathbf{O}^2, \dots, \mathbf{O}^L)$ are concatenated and linearly projected. The final global descriptor is L2-normalized to optimize it for subsequent similarity search.}
  \label{fig:arch}
\end{figure*}

Early visual place recognition (VPR) techniques relied on hand-crafted local features such as SIFT~\cite{lowe2004distinctive}, SURF~\cite{bay2006surf} and ORB~\cite{rublee2011orb}, which were aggregated into global descriptors using techniques like Bag-of-Words (BoW)~\cite{philbin2007object, torii2013visual, galvez2012bags} or Vector of Locally Aggregated Descriptors (VLAD)~\cite{jegou2011aggregating,arandjelovic2013all}.  BoW involves learning a \textit{visual vocabulary} (or set of clusters), where each \textit{visual word} (or cluster) represents a specific visual characteristic. Images are then represented as histograms, with bins indicating the frequency of each visual word. 
VLAD extends on this by capturing first-order statistics by accumulating the differences between local descriptors and their respective cluster centers.

\vspace{4pt}
\noindent\textbf{CNN-based VPR.}
The advent of deep learning marked a significant shift in VPR techniques, with various aggregation architectures proposed. Arandjelovic~\etal~\cite{arandjelovic2016netvlad} introduced NetVLAD, a trainable version of VLAD that integrates with CNN backbones, achieving superior performance over traditional methods. Following its success, many researchers proposed variants such as CRN~\cite{kim2017learned}, SPE-VLAD~\cite{yu2019spatial}, Gated~NetVLAD~\cite{zhang2021vector} and SSL-VLAD~\cite{nie2023training}. Other approaches focus on regions of interests within feature maps~\cite{babenko2015aggregating, tolias2015particular, zhang2023distilled}.
Another key aggregation method in image retrieval is the Generalized Mean (GeM)~\cite{radenovic2018fine}, which is a learnable form of global pooling. Building on GeM, Berton~\etal~\cite{zheng2021rethinking} have recently introduced CosPlace, enhancing GeM by integrating it with a linear projection layer, achieving remarkable performance for VPR. Ali-bey~\etal~\cite{ali2023mixvpr} introduced MixVPR, an all-MLP aggregation technique that achieved state-of-the-art performance on multiple benchmarks. Another facet of VPR is the training procedure, where the focus is on the loss function~\cite{ali2022gsv, Ali-Bey_2022_BMVC, berton2023eigenplaces, leyva2023data}.

\vspace{4pt}
\noindent\textbf{Transformer-based VPR.}
The Transformer architecture was initially introduced for natural language processing~\cite{vaswani2017attention}, and later adapted to Vision Transformers (ViT) for computer vision applications~\citep{dosovitskiy2020image}. In VPR, they have recently been used as backbone combined with MixVPR~\cite{huang2023dino, hou2023feature} or NetVLAD~\cite{9912433}  resulting in a performance boost compared to CNN backbones. Furthermore, AnyLoc~\cite{keetha2023anyloc} used features extracted from a foundation model, called DINOv2~\cite{oquab2023dinov2}, combined with unsupervised aggregation methods such as VLAD, resulting in notable performance on a wide range of benchmarks.

Recently, two-stage retrieval strategy has gained prominence in Transformers-based VPR. It starts with global retrieval using global descriptors to identify top candidates, followed by a computationally-intensive reranking phase that refines the results based on local features. In this context, Wang~\etal~\cite{wang2022transvpr} introduced TransVPR, which uses CNNs for feature extraction and Transformers for attention fusion to create global image descriptors, with additional patch-level descriptors for geometric verification.
On the other hand, Zhang~\etal~\cite{Zhang_2023_WACV} proposed ETR, a Transformer-based reranking technique that uses a CNN backbone for local and global descriptors. Their method leverages cross-attention between the local features of two images to cast the reranking as classification. Recently, Zhu~\etal~\cite{zhu2023r2former} proposed a unified framework integrating global retrieval and reranking, solely using transformers. For global retrieval, it employs the class token and utilizes other image tokens as local features for the reranking module. These two-stage techniques showed great performance, but at the expense of more computation and memory overheads. However, their global retrieval performance is still very limited compared to one-stage methods such as MixVPR~\cite{ali2023mixvpr} and CosPlace~\cite{berton2022rethinking}.
In this work, we propose, Bag-of-Queries (BoQ), a Transformer-based aggregation technique for global retrieval, that outperforms existing state-of-the-art without relying on reranking. This makes BoQ particularly suitable for applications where computational resources are limited, yet high accuracy and efficiency are essential.

\section{Methodology}
\label{sec:method}

In visual place recognition, effective aggregation of local features is crucial for accurate and fast global retrieval.
To address this, we propose the \textit{Bag-of-Queries} (BoQ) technique, a novel aggregation architecture that is end-to-end trainable and surprisingly simple, as depicted in \cref{fig:arch}.

Our technique rely on Multi-Head Attention (MHA) mechanism~\cite{vaswani2017attention}, which takes three inputs, queries ($\mathbf{q}$), keys ($\mathbf{k}$) and values ($\mathbf{v}$), and linearly project them into multiple parallel heads. The output, for each head, is computed as follows:
\begin{equation}
    \text{MHA}(\mathbf{q}, \mathbf{k}, \mathbf{v}) = \text{softmax} (\frac{\mathbf{q} \cdot \mathbf{k}^T}{\sqrt{d}}) \mathbf{v}.
\end{equation}
In this mechanism, the queries $\mathbf{q}$ play a crucial role. They act as a set of filters, determining which parts of the input (represented by keys $\mathbf{k}$ and values $\mathbf{v}$) are most relevant. The attention scores, derived from the dot product between queries and keys, essentially indicate to the model the degree of \emph{attention} to give to each part of the input. We refer to self-attention when $\mathbf{q}$, $\mathbf{k}$ and $\mathbf{v}$ are derived from the same input, \eg MHA($\mathbf{x}$, $\mathbf{x}$, $\mathbf{x}$). In contrast, cross-attention is the scenario where the query comes from a different source than the key and value.

Given an input image \(I \in \mathbb{R}^{3{\times}H{\times}W}\), we first process it through a backbone network, typically a pre-trained Convolutional Neural Network (CNN) or Vision Transformer (ViT), to extract its high-level features. For CNN backbones, we apply a $3{\times}3$ convolution to reduce their dimensionality, whereas, in the case of a ViT backbone, we apply a linear projection for the same purpose. We regard the result as a sequence of $N$ local features of dimension $d$, such as: \( \mathbf{X}^0 = [\mathbf{x}^0_1, \mathbf{x}^0_2, \ldots, \mathbf{x}^0_N] \). 
We then process \(\mathbf{X}^0\) through a sequence of Transformer-Encoder units and BoQ blocks. Each encoder transforms its input features, and passes the result to its subsequent BoQ block as follows:
\begin{equation}
\mathbf{X}^i = \text{Encoder}^i(\mathbf{X}^{i-1}).
\end{equation}
Here, $\mathbf{X}^{i-1}$ denotes the feature input to the $i$-th Encoder unit, and $\mathbf{X}^{i}$ represents the transformed output, which becomes the input for the next block in the pipeline. 

Each BoQ block contains a fixed set (\textit{a bag}) of \(M\) learnable queries, denoted as \( \mathbf{Q}^i = [\mathbf{q}^i_1, \mathbf{q}^i_2, \ldots, \mathbf{q}^i_M] \). These queries are trainable parameters of the model, independent of the input features (\eg, \verb|nn.Parameter| in PyTorch code), not to be confused with the term \say{\textit{query images}}, which refers to the test images used for benchmarking.

Before using $\mathbf{Q}^i$ to aggregate information in the $i^{th}$ BoQ block, we first apply self-attention between them:
\begin{equation}
\mathbf{Q}^i = \text{MHA}(\mathbf{Q}^i, \mathbf{Q}^i, \mathbf{Q}^i) + \mathbf{Q}^i .
\end{equation}
The self-attention operation allows the learnable queries to integrate their shared information during the training phase.
Next, we apply a cross-attention between \(\mathbf{Q}^i\) and and the input features $\mathbf{X}^i$ of the current BoQ block:
\begin{equation}
\mathbf{O}^i = \text{MHA}(\mathbf{Q}^i, \mathbf{X}^{i}, \mathbf{X}^{i}).
\end{equation}
This allows the learnable queries to dynamically assess the importance of each input feature by computing the relevance scores (attention weights) between $\mathbf{Q}^i$ and $\mathbf{X}^i$ and aggregate them into the output $\mathbf{O}^i$.

Following this, the outputs from each BoQ block are concatenated to form a single output.
\begin{equation}
\mathbf{O} = \text{Concat}(\mathbf{O}^1, \mathbf{O}^2, \dots, \mathbf{O}^L).
\end{equation}
This ensures that the final descriptor $\mathbf{O}$ combines information aggregated at different levels of the network, forming a rich and hierarchical representation.
The final step involves reducing the dimensionality of the output $\mathbf{O}$ using one or two successive linear projections with weights  \(\mathbf{W}_1\) and \(\mathbf{W}_2\), similar to the approach in~\cite{ali2022gsv, ali2023mixvpr}.
\begin{equation}
\text{Output} = \mathbf{W}_2 (\mathbf{W}_1 \mathbf{O})^T.
\end{equation}
This is followed by $L_2$-normalization, to bring the global descriptor to the unit hypersphere, which optimizes the similarity search. The training is performed using pair-based (or triplet-based) loss functions that are widely used in the literature~\cite{ali2022gsv}.

\vspace{4pt}
\noindent\textbf{Relations to other methods.} 
The Bag-of-Queries (BoQ) approach bears conceptual resemblance to Detection Transformer (DETR) model~\cite{carion2020end} in its use of object queries. However, BoQ is fundamentally different, as its global queries are exclusively used to extract and aggregate local feature information from the input; these queries do not contribute directly to the final representation, as demonstrated by the absence of any residual connection between the global queries and the output of the cross-attention. This is in contrast to DETR, where the object queries are integral part to the output, upon which object detection and classification is done.

In comparison to NetVLAD~\cite{arandjelovic2016netvlad}, which employs a set of learned cluster centers, and aggregates local features by calculating the weighted sum of their distances to each center. BoQ, leverages cross-attention between its learned queries and the input local feature to dynamically assess their relevance.

\section{Experiments}
\label{sec:experiments}
In this section, we conduct comprehensive experiments to demonstrate the effectiveness of our technique, Bag-of-Queries (BoQ), compared to current state-of-the-art methods. First, we describe the datasets used, and then the implementation details. This is followed by the evaluation metrics. We then provide a detailed comparative analysis of performance, ablation studies and qualitative results.

\subsection{Datasets}
Our experiments are performed on existing benchmarks that present a wide range of real-world challenges for VPR systems. \cref{tab:bench} provides a brief summary of these benchmarks. MapillarySLS (MSLS)~\cite{warburg2020mapillary} was collected using dashcams and features a broad range of viewpoints and lighting changes, testing the system's adaptability to typical variations in ground-level VPR. Pitts250k~\cite{torii2013visual} and Pitts30k datasets, extracted from Google Street View, exhibit significant changes in viewpoint, which tests the system's ability to maintain recognition across varying angles and perspectives. AmsterTime~\cite{yildiz2022amstertime} presents a unique challenge with historical grayscale images as queries and contemporary color images as references, covering temporal variations over several decades. Eynsham~\cite{cummins2009highly} consists entirely of grayscale images, adding complexity due to the absence of color information. Nordland~\cite{sunderhauf2013we} dataset was collected over four seasons, using a camera mounted on the front of a train. It encompasses scenes ranging from snowy winters to sunny summers, presenting the challenge of extreme appearance changes due to seasonal variations. Note that two variants of Nordland are used in VPR benchmarking: one uses a subset of 2760 summer images as queries against the entire winter sequence as reference images (marked with~*), while the other uses the entire winter sequence as query images against the entire summer sequence as reference images (marked with~**). SVOX~\cite{Berton_2021_svox} stands out with its comprehensive coverage of weather conditions, including overcast, rainy, snowy, and sunny images, to test adaptability to meteorological changes. Additionally, SVOX Night subset focuses on nocturnal scenes, a challenging scenario for most VPR systems.

\begin{table}[]
\centering
\resizebox{\columnwidth}{!}{%
\begin{tabular}{@{}lccccc@{}}
\toprule
                                       & \multicolumn{1}{l}{}           &               & \multicolumn{3}{c}{Variations}                                                             \\ \cmidrule(l){4-6} 
Dataset name                           & \multicolumn{1}{l}{\# quer.} & \# ref. & Viewpoint                    & Season                       & Day/Night                    \\ \midrule
MSLS~\cite{warburg2020mapillary}       & $740$                            & $18.9$k         & ${\times}$                   & ${\times} {\times}$          & ${\times}$                   \\
Pitts250k~\cite{torii2013visual}       & $8.2$k                           & $84$k           & ${\times} {\times} {\times}$ &                              &                              \\
Pitts30k~\cite{torii2013visual}        & $6.8$k                           & $10$k           & ${\times} {\times} {\times}$ &                              &                              \\
AmsterTime~\cite{yildiz2022amstertime} & $1231$                           & $1231$          & ${\times} {\times} {\times}$ & ${\times} {\times}$          & ${\times}$                   \\
Eynsham~\cite{cummins2009highly}       & $24$k                            & $24$k           & ${\times}$                   &                              &                              \\
Nordland*~\cite{zaffar2021vpr}          & $2760$                           & $27.6$k         &                              & ${\times} {\times} {\times}$ &                              \\
Nordland**~\cite{sunderhauf2013we}       & $27.6$k                          & $27.6$k         &                              & ${\times} {\times} {\times}$ &                              \\
St-Lucia~\cite{Milford_2008_st_lucia}  & $1464$                           & $1549$          &                              &                              &                              \\
SVOX~\cite{Berton_2021_svox}           & $14.3$k                          & $17.2$k         & ${\times}$                   & ${\times} {\times} {\times}$ & ${\times} {\times} {\times}$ \\
SPED~\cite{zaffar2021vpr}              & $607$                            & $607$           &                    & ${\times} {\times}$ & ${\times} {\times} $ \\ \bottomrule
\end{tabular}%
}
\caption{Overview of VPR benchmarks used in our experiments, highlighting their number of query images (\# quer.) and reference images (\# ref.), as well as their variations in terms of viewpoint, season, and day/night. The number of $\times$ marks indicates the degree of variations present (${\times}\text{ means low}$, ${\times}{\times}{\times}\text{ means high}$).}
\label{tab:bench}
\end{table}

\subsection{Implementation details}
\noindent\textbf{Architecture.} For our experiments, we primarily employ pre-trained ResNet-50 backbones for feature extraction~\cite{he2016deep}. Importantly, 
our proposed Bag-of-Queries (BoQ) technique is architecturally agnostic and seamlessly integrates with various CNNs or Vision Transformer (ViT) backbones. When employing a ResNet, we crop it at the second last residual to preserve a higher resolution of local features. Additionally, we freeze all but the last residual block and add a $3{\times}3$ convolution to reduce the number of channels, which decreases the computational and memory footprint of the self-attention mechanism. Implementing the BoQ model is straightforward with popular deep learning frameworks such as PyTorch~\cite{paszke2019pytorch} and TensorFlow~\cite{abadi2016tensorflow}, both providing highly optimized implementations of self-attention and cross-attention, which are the basic blocks of our technique.

\vspace{3pt}
\noindent\textbf{Training.} We train BoQ following the standard framework of GSV-Cities~\cite{ali2022gsv}, which provides a highly accurate dataset of $560$k images depicting $67$k places. For the loss function, we use Multi-Similarity loss~\cite{wang2019multi}, as it has been shown to perform best for visual place recognition~\cite{ali2022gsv}. We use batches containing $120-200$ places, each depicted by $4$ images resulting in mini-batches of $480-800$ images. We use the AdamW optimizer with a learning rate of $0.0002-0.0004$ depending the batch-size, and a weight decay of $0.001$. The learning rate is multiplied by $0.3$ after each $5-10$ epochs. Finally, we train for a maximum of $40$ epochs (including $3$ epochs of linear warmup) using images resized to $320 {\times} 320$. Note that while current techniques such as NetVLAD~\cite{arandjelovic2016netvlad}, CosPlace~\cite{zheng2021rethinking} and EigenPlaces~\cite{berton2023eigenplaces} train on images of size $480{\times}640$---effectively triple the pixel count of our chosen resolution---our BoQ model still achieves better performance, underscoring its effectiveness.

\vspace{3pt}
\noindent\textbf{Evaluation metrics.} We follow the same evaluation metric of existing literature \cite{arandjelovic2016netvlad, kim2017learned, warburg2020mapillary, zaffar2021vpr, wang2022transvpr, berton2022rethinking, ali2023mixvpr, ali2022gsv, Ali-Bey_2022_BMVC, berton2023eigenplaces}, where the recall@$k$ is measured. Recall@$k$ is defined as the percentage of query images for which at least one of the top-$k$ predicted reference images falls within a predefined threshold distance, which is typically $25$ meters. However, for the Nordland benchmark, which features aligned sequences, the threshold is $10$ frame counts for Nordland** and only $1$ frame for Nordland*.

\subsection{Comparison with previous works}
\setlength{\tabcolsep}{10pt}
\begin{table*}[]
\centering
\resizebox{\textwidth}{!}{%
\begin{tabular}{@{}l c c@{\hspace{5pt}}c@{\hspace{5pt}}c c@{\hspace{5pt}}c@{\hspace{5pt}}c c@{\hspace{5pt}}c@{\hspace{5pt}}c c@{\hspace{5pt}}c@{\hspace{5pt}}c@{}}
\toprule
\multirow{2}{*}{Method}  &  \multirow{2}{*}{Dim.}               & \multicolumn{3}{c}{Pitts250k-test}                  & \multicolumn{3}{c}{MSLS-val}                        & \multicolumn{3}{c}{SPED}                            & \multicolumn{3}{c}{Nordland*}                       \\ 
\cmidrule(l{0.3cm} r{0.3cm}){3-5} \cmidrule(l{0.3cm} r{0.3cm}){6-8} \cmidrule(l{0.3cm} r{0.3cm}){9-11} \cmidrule(l{0.3cm} r{0.0cm}){12-14} 
                          &               & R@1             & R@5             & R@10            & R@1             & R@5             & R@10            & R@1             & R@5             & R@10            & R@1             & R@5             & R@10            \\ \midrule
AVG~\cite{arandjelovic2016netvlad}     & $2048$ & $78.3$          & $89.8$          & $92.6$          & $73.5$          & $83.9$          & $85.8$          & $58.8$          & $77.3$          & $82.7$          & $15.3$          & $27.4$          & $33.9$          \\
GeM \cite{radenovic2018fine}           & $2048$ & $82.9$          & $92.1$          & $94.3$          & $76.5$          & $85.7$          & $88.2$          & $64.6$          & $79.4$          & $83.5$          & $20.8$          & $33.3$          & $40.0$          \\
NetVLAD \cite{arandjelovic2016netvlad} & $32768$ & $90.5$          & $96.2$          & $97.4$          & $82.6$          & $89.6$          & $92.0$          & $78.7$          & $88.3$          & $91.4$          & $32.6$          & $47.1$          & $53.3$          \\
SPE-NetVLAD \cite{yu2019spatial}       & $163840$ & $89.2$          & $95.3$          & $97.0$          & $78.2$          & $86.8$          & $88.8$          & $73.1$          & $85.5$          & $88.7$          & $25.5$          & $40.1$          & $46.1$          \\
Gated NetVLAD \cite{zhang2021vector}   & $32768$ & $89.7$          & $95.9$          & $97.1$          & $82.0$          & $88.9$          & $91.4$          & $75.6$          & $87.1$          & $90.8$          & $34.4$          & $50.4$          & $57.7$          \\
Conv-AP \cite{ali2022gsv}              & $4096$ & $92.4$          & $97.4$          & $98.4$          & $83.4$          & $90.5$          & $92.3$          & $80.1$          & $90.3$          & $93.6$          & $38.2$          & $54.8$          & $61.2$          \\
CosPlace \cite{berton2022rethinking}   & $2048$ & $92.3$          & $97.4$          & $98.4$          & $87.4$          & $\underline{93.8}$          & $94.9$          & $75.3$          & $85.9$          & $88.6$          & $54.4$          & $69.8$          & $75.9$          \\
MixVPR \cite{ali2023mixvpr}        &  $4096$    & $\underline{94.2}$          & $\underline{98.2}$          & $\underline{98.9}$          & $88.0$          & $92.7$          & $94.6$          & $\underline{85.2}$          & $\underline{92.1}$          & $94.6$          & $\underline{58.4}$          & $\underline{74.6}$          & $\underline{80.0}$          \\
EigenPlaces \cite{berton2023eigenplaces}& $2048$ & $94.1$          & $97.9$          & $98.7$          & $\underline{89.2}$          & $93.6$          & $\underline{95.0}$          & $82.4$          & $91.4$          & $\underline{94.7}$          & $54.2$          & $68.0$          & $73.9$          \\ \midrule
$\mathbf{BoQ}$ \textbf{(Ours)}         &  $4096$   & $\mathbf{95.0}$ & $98.4$ & $\mathbf{99.1}$ & $91.1$          & $94.8$ & $95.7$            & $85.4$          & $93.1$          & $95.4$          & $69.5$          & $83.4$          & $87.0$ \\ 
$\mathbf{BoQ}$ \textbf{(Ours)}         &  $16384$  & $\mathbf{95.0}$ & $\mathbf{98.5}$          & $\mathbf{99.1}$ & $\mathbf{91.2}$ & $\mathbf{95.3}$          & $\mathbf{96.1}$ & $\mathbf{86.5}$   & $\mathbf{93.4}$ & $\mathbf{95.7}$ & $\mathbf{70.7}$ & $\mathbf{84.0}$ & $\mathbf{87.5}$ \\ \bottomrule
\end{tabular}%
}
\caption{Comparison of our technique, Bag-of-Queries ($\mathit{BoQ}$), to existing state-of-the-art methods. Best results are in shown in \textbf{bold} and second best are \underline{underlined}. All methods use a pre-trained ResNet-50 backbone and follow identical training procedures on the GSV-Cities dataset, except for CosPlace and EigenPlaces where authors' pre-trained weights demonstrated superior performance and are thus used here.}
\label{tab:sota_1}
\end{table*}

\setlength{\tabcolsep}{5pt}
\begin{table*}[]
\centering
\resizebox{\textwidth}{!}{%
\begin{tabular}{@{}lccccccccccc@{}}
\toprule
            &           & \multicolumn{3}{c}{Multi-view datasets}    & \multicolumn{7}{c}{Frontal-view datasets}                                                                                                                                                                                                                                                                       \\ \cmidrule(l{0.3cm} r{0.3cm}){3-5}  \cmidrule(l{0.3cm} r{0.0cm}){6-12}
Method      & Backbone  & AmsterTime & Eynsham       & Pitts30k      & Nordland**    & St Lucia      & \begin{tabular}[c]{@{}c@{}}SVOX\\ Night\end{tabular} & \begin{tabular}[c]{@{}c@{}}SVOX\\ Overcast\end{tabular} & \begin{tabular}[c]{@{}c@{}}SVOX\\ Rain\end{tabular} & \begin{tabular}[c]{@{}c@{}}SVOX\\ Snow\end{tabular} & \begin{tabular}[c]{@{}c@{}}SVOX\\ Sun\end{tabular} \\ \midrule
NetVLAD \cite{arandjelovic2016netvlad}    & VGG-16    & $16.3$       & $77.7$          & $85.0$          & $13.1$          & $64.6$          & $8.0$                                                  & $66.4$                                                    & $51.5$                                                & $54.4$                                                & $35.4$                                               \\
SFRS \cite{ge2020self}        & VGG-16    & $29.7$       & $72.3$          & $89.1$          & $16.0$          & $75.9$          & $28.6$                                                 & $81.1$                                                    & $69.7$                                                & $76.0$                                                & $54.8$                                               \\
CosPlace \cite{berton2022rethinking}   & VGG-16    & $38.7$       & $88.3$          & $88.4$          & $58.5$          & $95.3$          & $44.8$                                                 & $88.5$                                                    & $85.2$                                                & $89.0$                                                & $67.3$                                               \\
EigenPlaces \cite{berton2023eigenplaces} & VGG-16    & $38.0$       & $89.4$          & $89.7$          & $54.5$          & $95.4$          & $42.3$                                                 & $89.4$                                                    & $83.5$                                                & $89.2$                                                & $69.7$                                               \\ \midrule
Conv-AP \cite{ali2022gsv}    & ResNet-50 & $33.9$       & $87.5$          & $90.5$          & $62.9$          & {\ul $99.7$}    & $43.4$                                                 & $91.9$                                                    & $82.8$                                                & $91.0$                                                & $80.4$                                               \\
CosPlace \cite{berton2022rethinking}   & ResNet-50 & $47.7$       & $90.0$          & $90.9$          & $71.9$          & $99.6$          & $50.7$                                                 & $92.2$                                                    & $87.0$                                                & $92.0$                                                & $78.5$                                               \\
MixVPR \cite{ali2023mixvpr}      & ResNet-50 & $40.2$       & $89.4$          & $91.5$          & {\ul $76.2$}    & $99.6$          & {\ul $64.4$}                                           & {\ul $96.2$}                                              & {\ul $91.5$}                                          & {\ul $96.8$}                                          & $84.8$                                               \\
EigenPlaces \cite{berton2023eigenplaces} & ResNet-50 & {\ul $48.9$}       & {\ul $90.7$}    & $\mathbf{92.5}$ & $71.2$          & $99.6$          & $58.9$                                                 & $93.1$                                                    & $90.0$                                                & $93.1$                                                & {\ul $86.4$}                                         \\
$\mathbf{BoQ}$ \textbf{(Ours)}  & ResNet-50 & $\mathbf{52.2}$       & $\mathbf{91.3}$ & {\ul 92.4}    & $\mathbf{83.1}$ & $\mathbf{100}$ & $\mathbf{87.1}$                                        & $\mathbf{97.8}$                                           & $\mathbf{96.2}$                                       & $\mathbf{98.7}$                                       & $\mathbf{95.9}$                                      \\ \bottomrule
\end{tabular}%
}
\caption{Recall@1 comparison across multi-view and frontal-view datasets for various techniques, including our Bag-of-Queries (BoQ). Our method achieves state-of-the-art performance on most benchmarks, demonstrating robustness to extreme weather and illumination changes, particularly on the challenging SVOX dataset.}
\label{tab:sota_2}
\end{table*}

\begin{table}[]
\centering
\resizebox{\columnwidth}{!}{%
\begin{tabular}{@{}lcccccc@{}}
\toprule
                                         & \multirow{2}{*}{\begin{tabular}[c]{@{}c@{}}Extrac.\\ (ms)\end{tabular}} & \multirow{2}{*}{\begin{tabular}[c]{@{}c@{}}Rerank.\\ (ms)\end{tabular}} & \multicolumn{2}{c}{MSLS-val}  & \multicolumn{2}{c}{Pitts30k}  \\ 
                                         \cmidrule(l{0.1cm} r{0.1cm}){4-5} \cmidrule(l{0.1cm} r{0.0cm}){6-7}
                                         &                                                                            &                                                                           & R@1           & R@5           & R@1           & R@5           \\ \midrule
SP-SuperGlue \cite{detone2018superpoint} &                     $160$                                                       &                      $7500$                                                     & $78.1$          & $81.9$          & $87.2$          & $94.8$          \\
DELG \cite{cao2020unifying}              &                     $190$                                                       &                      $35200$                                                     & $83.2$          & $90.0$            & $89.9$          & $95.4$          \\
Patch-NetVLAD \cite{hausler2021patch}    &                     $1300$                                                       &                     $7400$                                                     & $79.5$          & $86.2$          & $88.7$          & $94.5$          \\
TransVPR \cite{wang2022transvpr}         &                     $45$                                                       &                      $3200$                                                     & $86.8$          & $91.2$          & $89.0$            & $94.9$          \\
R2Former \cite{zhu2023r2former}          &                     $31$                                                       &                      $400$                                                     & $89.7$          & $\mathbf{95.0}$            & $91.1$          & $95.2$          \\ \midrule
$\mathbf{BoQ}$ \textbf{(Ours)}           &                     $7$                                                       &                      $0$                                                     & $\mathbf{91.4}$ & $94.5$ & $\mathbf{92.4}$ & $\mathbf{96.2}$ \\ \bottomrule
\end{tabular}%
}
\caption{\textbf{Comparing against two-stage retrieval techniques} in terms of performance and latency of feature extraction and reranking (when applicable). The first $5$ techniques use a second refinement pass (geometric matching) to re-rank the top $100$ candidates in order to improve retrieval performance. BoQ (ours) does not employ re-ranking, which makes it orders of magnitude faster than the fastest two-stage technique.}
\label{tab:boq_vs_reranking}
\end{table}

In this section, we present an extensive comparison of our proposed method, Bag-of-Queries (BoQ), with a wide range of existing state-of-the-art VPR aggregation techniques. This includes global average pooling (AVG)~\cite{arandjelovic2016netvlad}, Generalized Mean~(GeM)~\cite{radenovic2018fine},  NetVLAD~\cite{arandjelovic2016netvlad} alongside its recent variants, SPE-NetVLAD~\cite{yu2019spatial} and Gated~NetVLAD~\cite{zhang2021vector}. Importantly, we compare against recent cutting-edge techniques, such as ConvAP~\cite{ali2022gsv}, CosPlace~\cite{zheng2021rethinking}, EigenPlaces~\cite{berton2023eigenplaces} and MixVPR~\cite{ali2023mixvpr}, which currently hold the best scores in most existing benchmarks, setting a high standard for BoQ to measure up against in the field of VPR.

Moreover, despite BoQ being a global retrieval technique which does not employ reranking, we compare it with current state-of-the-art two-stage retrieval techniques~\cite{detone2018superpoint, cao2020unifying, hausler2021patch, wang2022transvpr, zhu2023r2former}. These methods leverage geometric verification to enhance recall@k performance at the expense of increased computation and memory overhead.

\vspace{3pt}
\noindent\textbf{Results analysis.}
In \cref{tab:sota_1}, we present a comparative analysis across several challenging datasets, focusing on Recall@1 (R@1), Recall@5 (R@5), and Recall@10 (R@10). We also offer insight into the performance of each method with regard to the top-ranked retrieved images.
\begin{itemize}
    \item On Pitts250k-test, BoQ outperforms all other methods with a marginal yet notable improvement as it is the first global technique that reaches the $95\%$ R@1 threshold on this benchmark.
    \item On MSLS-val and SPED, BoQ outperforms the second best methods (EigenPlaces and MixVPR) by $2.0$ and $1.3$ percentage points respectively.
    \item On the challenging Nordland* benchmark, known for its extreme seasonal variations, BoQ significantly outperforms all compared methods, indicating its robustness to severe environmental changes. Surpassing CosPlace and Eigenplaces by $16$, and MixVPR by $12$ percentage points, which translates to over $400$ additional accurate retrievals within the top-$1$ results.
\end{itemize}

\noindent In \cref{tab:sota_2}, we follow the same evaluation style proposed by Berton~\etal~\cite{berton2023eigenplaces}, where benchmarks are categorized into multi-view datasets, where images are oriented in various angles, and frontal-view dataset where images are mostly forward facing. Top-1 retrieval performance (R@1) is reported for each technique.
\begin{itemize}
    \item For the multi-view benchmarks, BoQ achieves highest R@1 on AmsterTime and Eynsham, and second best compared to EigenPlaces on Pitts30k-test. Both techniques show robustness to viewpoint changes in urban environments, while facing challenges with the AmsterTime dataset, which includes decades-old historical imagery. 
    \item On frontal-view benchmarks, particularly on Nordland** (which comprises $27.6$k queries), BoQ achieves R@1 of $83.1\%$, which is $7$ and $11$ percentage points more than MixVPR, and CosPlace respectively. This translates to BoQ correctly retrieving an additional $2400$ and $3700$ images within the top-1 results.
    \item On SVOX, BoQ achieves best results under all conditions, especially on SVOX night, where it achieves $87.1\%$ R@1, outperforming the second best method (MixVPR) by $22.7$ percentage points. This highlights BoQ's potential under low-light conditions.
\end{itemize}

\noindent Furthermore, we compare BoQ to two-stage retrieval techniques in \cref{tab:boq_vs_reranking}, by reporting performance on MSLS-val and Pitts30k-test. Although our technique does not perform reranking, its global retrieval performance surpasses that of existing two-stage techniques, including Patch-NetVLAD (2021), TransVPR (2022) and R2Former (2023). Importantly, this makes our approach more efficient in terms of memory and computation: over $30{\times}$ faster retrieval time compared to R2Former.

\subsection{Ablation studies}
\noindent\textbf{Number of learnable queries.} 
We conduct an ablation study to investigate the impact of varying the number of learnable queries, \(\mathbf{Q}\), on the performance of BoQ. We use a ResNet-50 backbone and two BoQ blocks. The results are presented in \cref{tab:nb_q}. We observe that performance improves as the number of queries increases. However, the benefits of additional queries vary across datasets. In less diverse environments, such as Pitts30k, adding more queries yields marginal performance improvements  (increasing from $8$ to $64$ queries results in a mere $0.2\%$ R@1 performance gain). In contrast, the results on AmsterTime, which features highly diverse images spanning decades, demonstrate that the model benefits significantly from additional queries. This aligns with the underlying intuition of BoQ, where each query implicitly learns a distinct universal pattern.
\setlength{\tabcolsep}{10pt}
\begin{table}[]
\centering
\resizebox{\columnwidth}{!}{%
\begin{tabular}{@{}ccccccc@{}}
\toprule
\multirow{2}{*}{\begin{tabular}[c]{@{}c@{}}Size of \\ $\mathbf{Q}$\end{tabular}} & \multicolumn{2}{c}{MSLS-val} & \multicolumn{2}{c}{Pitts30k-val} & \multicolumn{2}{c}{AmsterTime} \\ 
\cmidrule(l{0.3cm} r{0.3cm}){2-3} \cmidrule(l{0.3cm} r{0.3cm}){4-5} \cmidrule(l{0.3cm} r{0.0cm}){6-7}
                                                                          & R@1           & R@5          & R@1             & R@5            & R@1         & R@5        \\ \midrule
$4$                                                                         & $86.9$          & $92.7$         & $93.1$            & $97.9$           & $42.7$        & $61.4$       \\
$8$                                                                         & $88.1$          & $93.4$         & $93.9$            & $98.4$           & $44.3$        & $63.8$       \\
$16$                                                                        & $88.7$          & $93.4$         & $94.0$            & $98.5$           & $46.2$        & $68.4$       \\
$32$                                                                        & $90.6$          & $94.0$         & $94.1$            & $98.6$           & $48.9$        & $69.0$       \\
$64$                                                                        & $91.3$          & $94.3$         & $94.5$            & $98.9$           & $52.0$          & $70.1$       \\ \bottomrule
\end{tabular}%
}
\caption{\textbf{Ablation on the number of learnable queries}. We vary the size of $\mathbf{Q}$ in each BoQ block. ResNet-50 is used as backbone, with two BoQ blocks. Performance increases with the number of queries until stagnation at $64$ queries.}
\label{tab:nb_q}
\end{table}

\setlength{\tabcolsep}{5pt}
\begin{table}[]
\centering
\resizebox{\columnwidth}{!}{%
\begin{tabular}{@{}cccccc@{}}
\toprule
\begin{tabular}[c]{@{}c@{}}\# blocks\\ ($\times L$)\end{tabular} & \begin{tabular}[c]{@{}c@{}}\# params \\ (M)\end{tabular} & MSLS & Nordland & AmsterTime & \begin{tabular}[c]{@{}c@{}}SVOX \\ Night\end{tabular} \\ \midrule
$1$                                                                & $2.5 $                                                     & $87.1$     & $56.7$     & $40.5$       & $62.8$                                                  \\
$2$                                                                & $4.3 $                                                     & $88.0$     & $61.1$     & $42.2$       & $68.6$                                                  \\
$4$                                                                & $8.0 $                                                     & $88.1$     & $62.7$     & $44.0$       & $71.4$                                                  \\
$8$                                                                & $15.4$                                                     & $87.2$     & $59.4$     & $43.0$       & $70.1$                                                  \\ \bottomrule
\end{tabular}%
}
\caption{\textbf{Ablation on the number of BoQ blocks used.} Recall@1 performance is reported for each configuration. 
ResNet-18 is used as backbone. The total number of parameters (in millions) is reported.}
\label{tab:nb_boqs}
\end{table}

\begin{table}[]
\resizebox{\columnwidth}{!}{%
\begin{tabular}{@{}lcccc@{}}
\cmidrule(l){2-5}
                                & MSLS-val      & Pitts30k      & Pitts250k         & Nordland \\ \midrule
BoQ w/out self-att.             & $87.1$        & $90.7$        & $92.3$            & $56.4$     \\
\textbf{BoQ with self-att.}     & $\mathbf{88.4}$     & $\mathbf{91.5}$     & $\mathbf{93.3}$      & $\mathbf{65.9}$     \\ \bottomrule
\end{tabular}%
}
\caption{\textbf{Ablation on the use of self-attention} between the global learned queries ($\mathbf{Q}^i$). Recall@1 is reported. ResNet-18 is used as backbone.} 
\label{tab:sa_ablation}
\end{table}

\setlength{\tabcolsep}{3pt}

\begin{table}[]
\centering
\resizebox{\columnwidth}{!}{%
\begin{tabular}{@{}lccccccc@{}}
\toprule
\multirow{2}{*}{Backbone} & \multirow{2}{*}{\begin{tabular}[c]{@{}c@{}}\# params\\ (M)\end{tabular}} & \multicolumn{2}{c}{MSLS-val} & \multicolumn{2}{c}{SPED} & \multicolumn{2}{c}{Nordland*} \\ 
\cmidrule(l{0.1cm} r{0.1cm}){3-4}  \cmidrule(l{0.1cm} r{0.1cm}){5-6}  \cmidrule(l{0.1cm} r{0.0cm}){7-8} 
                          &                                                                          & R@1           & R@5          & R@1         & R@5        & R@1           & R@5           \\ \midrule
ResNet-18                 & $7.1$                                                                      & $88.4$          & $92.6$         & $83.5$        & $92.7$       & $65.9$          & $79.3$          \\
ResNet-34                 & $12.5$                                                                     & $89.5$          & $93.1$         & $85.5$        & $92.6$       & $64.1$          & $80.1$          \\
ResNet-50                 & $14.6$                                                                     & $91.4$          & $94.5$         & $86.2$        & $94.4$       & $74.4$          & $86.1$          \\
ResNet-101                & $42.9$                                                                     & $90.5$          & $95.4$         & $85.6$        & $93.8$       & $71.9$          & $84.9$          \\ \bottomrule
\end{tabular}%
}
\caption{\textbf{Ablation on the backbone architecture.} Each backbone is cropped at the second last residual block. Two BoQ blocks are used, with $64$ learnable queries in each. We show the total number of parameters. BoQ achieves state-of-the-art performance with only a ResNet-34 backbone, which highlights its potential use for real-time scenarios.}
\label{tab:backbones}
\end{table}

\vspace{4pt}
\noindent\textbf{Number of BoQ blocks.}
To assess the influence of the number of BoQ blocks within our architecture, we conduct experiments by varying $L$, the number BoQ blocks utilized. We use a ResNet-18 backbone followed by $L$ BoQ blocks, each comprising a bag of $32$ learnable queries. We report R@1 performance for each setup. The results, presented in \cref{tab:nb_boqs}, demonstrate that even when paired with a lightweight backbone like ResNet-18, BoQ remains competitive against state-of-the-art methods such as CosPlace~\cite{berton2022rethinking} and MixVPR~\cite{ali2023mixvpr}, which use ResNet-50 backbone. Best overall performance is observed when employing $4$ BoQ blocks.

\vspace{4pt}
\noindent\textbf{Ablation on the self-attention.}
Applying self-attention on ($\mathbf{Q}^i$) brings more stability/performance by adding context between the queries. As shown in \cref{tab:sa_ablation}, we train a BoQ model with ResNet-18 backbone for $35$ epochs. Adding self-attention between the learnable queries brings consistent performance improvement across various benchmarks. Note that the self-attention's output can be cached during eval/test to avoid recomputation at every test iteration.
\setlength{\tabcolsep}{8pt}

\vspace{4pt}
\noindent\textbf{Backbone architecture.}
In \cref{tab:backbones} we present a comparative performance of BoQ coupled with different ResNet backbone architectures. For each backbone, we crop it at the second last residual block and follow it with two BoQ blocks, each comprising $64$ learnable queries. The total number of parameters (trainable + non trainable) is provided. The empirical results showcase that when BoQ is coupled with ResNet-34, a relatively lightweight backbone, it achieves competitive performance compared to NetVLAD, CosPlace and EigenPlaces coupled with ResNet-50 backbone. Interestingly, using ResNet-101, a relatively deeper backbone, we did not achieve better performance than ResNet-50, which could be attributed to memory constraints necessitating a smaller training batch size.

\subsection{Vision Transformer Backbones}

All previous experiments in this paper were conducted using a ResNet~\cite{he2016deep} backbone to ensure fairness against existing state-of-the-art methods. Recently, vision foundation models such as DINOv2~\cite{oquab2023dinov2} have been introduced and rapidly adopted as the de facto backbone for various computer vision tasks. In this experiment, we evaluate the performance of BoQ when using a DINOv2 backbone. We employ the pre-trained base variant with 86M parameters, freezing all but the last two blocks to enable fine-tuning. We train on GSV-Cities~\cite{ali2022gsv} with images resized to $280{\times}280$ and perform testing with images resized to $322{\times}322$. The results presented in \cref{tab:dinov2}, demonstrate that DINOv2 further enhances the performance of BoQ, achieving new all-time high performance on all benchmarks.

\setlength{\tabcolsep}{6pt}
\begin{table}[]
\centering
\resizebox{\columnwidth}{!}{%
\begin{tabular}{@{}lccc@{\hspace{8mm}}ccc@{}}
\toprule
\multirow{2}{*}{\begin{tabular}[c]{@{}l@{}}Test\\ Dataset\end{tabular}} & \multicolumn{3}{c}{DinoV2-BoQ \phantom{xxxx}} & \multicolumn{3}{c}{ResNet50-BoQ} \\ 
\cmidrule(l{2mm} r{10mm}){2-4} \cmidrule(l{1mm} r{2mm}){5-7}
                                                                        & R@1      & R@5      & R@10     & R@1       & R@5       & R@10     \\ \midrule
MSLS-val~\cite{warburg2020mapillary}                                                                & $93.8$     & $96.8$     & $97.0$     & $91.2$      & $95.3$      & $96.1$     \\
Pitts30k-test~\cite{torii2013visual}                                                           & $93.7$     & $97.1$     & $97.9$     & $92.4$      & $96.5$      & $97.4$     \\
Pitts250k-test~\cite{torii2013visual}                                                          & $96.6$     & $99.1$     & $99.5$     & $95.0$      & $98.5$      & $99.1$     \\
Nordland*~\cite{zaffar2021vpr}                                                              & $81.3$     & $92.5$     & $94.8$     & $70.7$      & $84.0$      & $87.5$     \\
Nordland**~\cite{sunderhauf2013we}                                                               & $90.6$     & $96.0$     & $97.5$     & $83.1$      & $91.0$      & $93.5$     \\
SPED~\cite{zaffar2021vpr}                                                                    & $92.5$     & $95.9$     & $96.7$     & $86.5$      & $93.4$      & $95.7$     \\
AmesterTime~\cite{yildiz2022amstertime}                                                             & $63.0$     & $81.6$     & $85.1$     & $52.2$      & $72.5$      & $78.4$     \\
Eynsham~\cite{cummins2009highly}                                                                 & $92.2$     & $95.6$     & $96.4$     & $91.3$      & $94.9$      & $95.9$     \\
St-Lucia~\cite{Milford_2008_st_lucia}                                                                & $100$     & $100  $    & $100  $    & $100$       & $100$       & $100$      \\ \midrule
SVOX-night~\cite{Berton_2021_svox}                                                              & $97.7$     & $99.5$     & $99.8$     & $87.1$      & $94.4$      & $96.1$     \\
SVOX-sun~\cite{Berton_2021_svox}                                                                & $97.5$     & $99.4$     & $99.5$     & $95.9$      & $98.8$      & $98.9$     \\
SVOX-rain~\cite{Berton_2021_svox}                                                               & $98.8$     & $99.7$     & $99.8$     & $96.2$      & $99.5$      & $99.7$     \\
SVOX-snow~\cite{Berton_2021_svox}                                                               & $99.4$     & $99.7$     & $99.9$     & $98.7$      & $99.5$      & $99.7$     \\
SVOX-overcast~\cite{Berton_2021_svox}                                                           & $98.5$     & $99.3$     & $99.4$     & $97.8$      & $99.2$      & $99.3$     \\
SVOX-all~\cite{Berton_2021_svox}                                                                & $99.0$     & $99.6$     & $99.7$     & $98.9$      & $99.4$      & $99.6$     \\ \midrule
Tokyo 24/7~\cite{torii201524}                                                              & $98.1$     & $98.1$     & $98.7$     & $94.3$      & $96.5$      & $96.5$   \\ 
MSLS-Challenge~\cite{warburg2020mapillary}                                                          & $79.0$     & $90.3$     & $92.0$     & --          & --          & --       \\
SF-Landmark~\cite{chen2011city}                                                             & $93.6$     & $95.8$     & $96.5$     & --          & --          & --       \\ 
\bottomrule
\end{tabular}%
}
\caption{Performance comparison of BoQ using DINOv2 and ResNet-50 backbone.}
\label{tab:dinov2}
\end{table}

\subsection{Visualization of the learned queries}
\begin{figure}
\begin{center}
\includegraphics[width=1\linewidth]{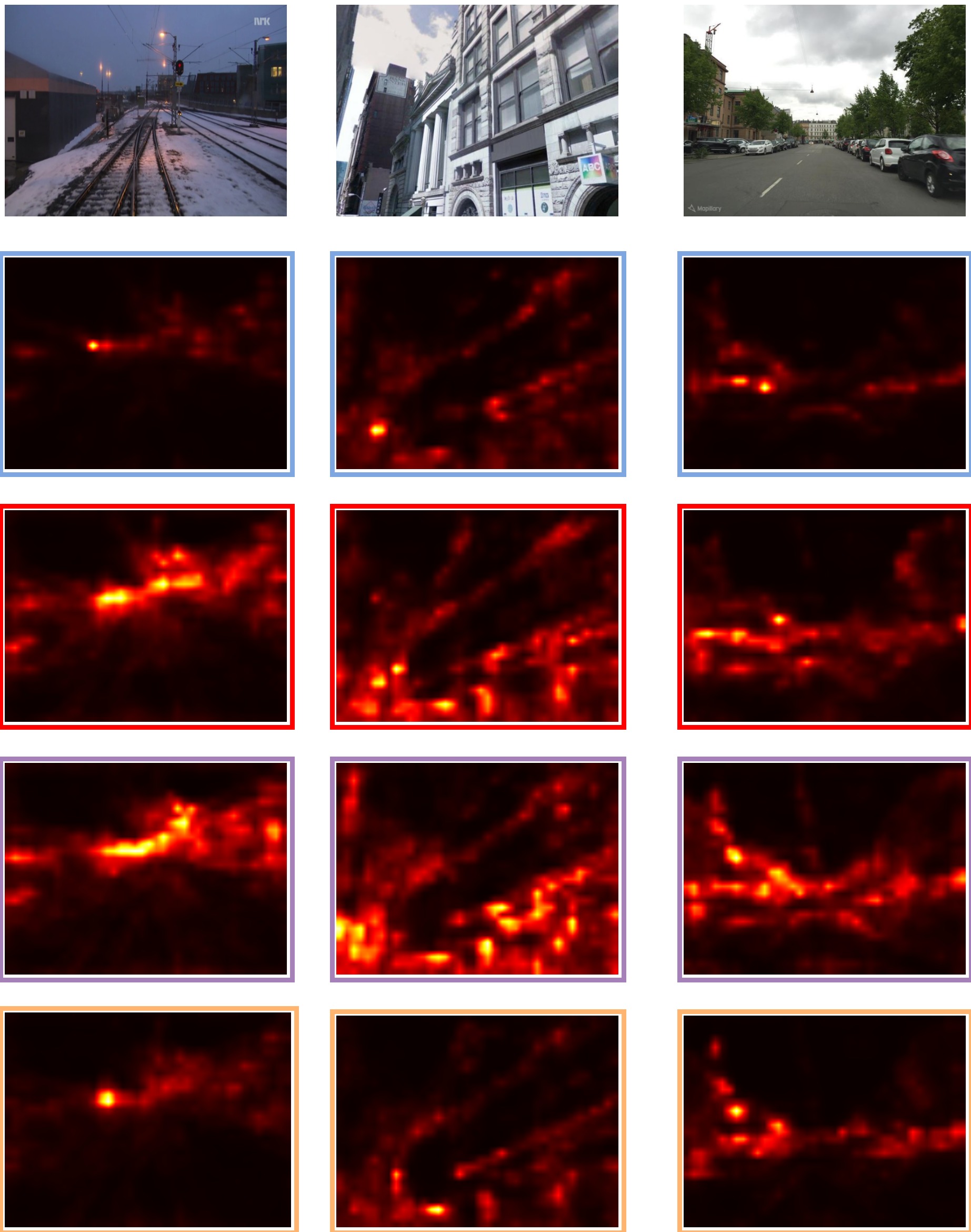}
\end{center}
   \caption{Visualization of the cross attention weights between the input images and the learned queries. The three examples are from Nordland, Pitts30k and MSLS datasets, respectively. 
   We selected four queries (among $64$) from the second BoQ block of a trained network. Vertically, we can see how the input image is aggregated by each query. The aggregation is done through the product of the weight with the input feature maps, resulting in one aggregated descriptor per query. Horizontally, we can see in each line how each query spans the input image. For example, the first query looks more for fine grained details, while the second looks more for large areas in the input images.}
\label{fig:learnable_q}
\end{figure}

\cref{fig:learnable_q} illustrates the cross-attention weights between the input image and a subset of learned queries within the BoQ blocks. We highlight four distinct queries (among $64$ in total) from the second BoQ block, in order to understand their individual aggregation characteristics. Observing vertically, we can see how the input image is seen through the \say{lenses} of each learned query. The aggregation is realized through the product of the cross-attention weights with the input feature maps, resulting in a single aggregated descriptor per query. Notice that the role of each query is to \say{scan}---via cross-attention---the input features and generate the aggregation weights.

Observing horizontally, the diversity in attention patterns across the learned queries becomes apparent. For instance, the first query (top row) appears to concentrate on fine-grained details within the feature maps, as indicated by the intense localized areas. In contrast, the second query (second row) shows a preference for larger regions in the input features, suggesting a more holistic capture of scene attributes.

\section{Conclusion}
\label{sec:sonclusion}
In this work, we introduced \emph{Bag-of-Queries (BoQ)}, a novel aggregation technique for visual place recognition, which is based on the use of learnable global queries to probe local features via a cross-attention mechanism, allowing for robust and discriminative feature aggregation.
Our extensive experiments on $14$ different large-scale benchmarks demonstrate that BoQ consistently outperforms current state-of-the-art techniques, particularly in handling complex variations in viewpoint, lighting, and seasonal changes. Furthermore, BoQ being a global (one-stage) retrieval technique, it outperforms existing two-stage retrieval methods that employ geometric verification for re-ranking, all while being orders of magnitude faster, setting a new standard for VPR research.

In future work, building upon the concepts discussed in \cref{sec:method}, the spatial information carried in the output of the last encoder, denoted as \(\mathbf{X}^L\), presents an opportunity for further enhancement through the application of special reranking strategies.

\vspace{5pt}
\noindent\textbf{Acknowledgement:} This work has been supported by The Fonds de Recherche du Québec Nature et technologies (FRQNT 254912). We gratefully acknowledge the support of NVIDIA Corporation with the donation of a Quadro RTX 8000 GPU used for our experiments.

{
    \small
    \bibliographystyle{ieeenat_fullname}
    \bibliography{all_references}
}

\clearpage
\setcounter{page}{1}
\maketitlesupplementary

\begin{figure*}
  \centering
  \includegraphics[width=1\linewidth]{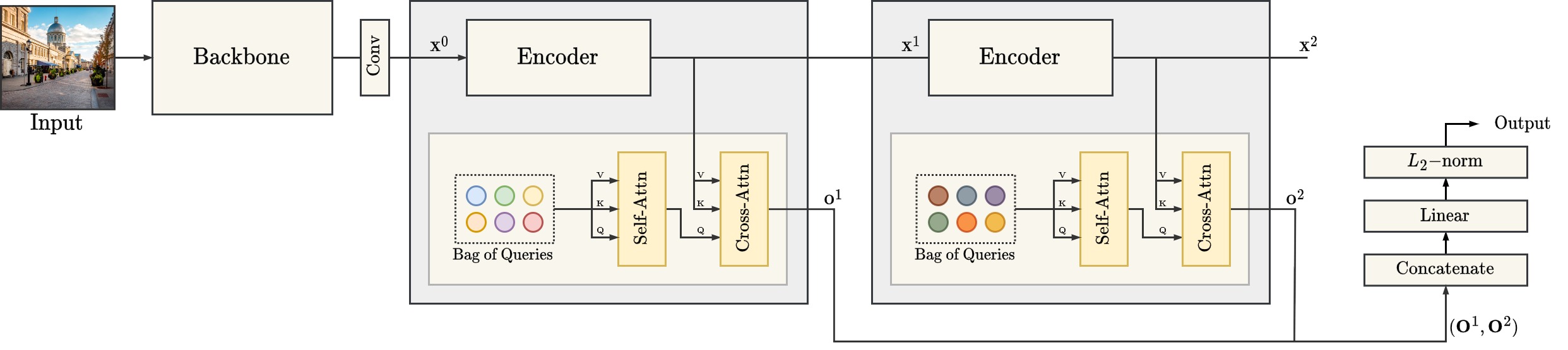}
  \caption{Detailed architecture of our model using ResNet-50 backbone and two BoQ blocks.}
  \label{fig:arch_2BOQ}
\end{figure*}

\section{More implementation details}
\label{sec:more_imp}
\noindent\textbf{Training image size.}
For our training implementation, we resized the images to $320{\times}320$ pixels to maintain consistency with the training procedures of \cite{ali2023mixvpr, ali2022gsv, Ali-Bey_2022_BMVC}. This choice of specific resolution directly influences the spatial resolution of the resulting feature maps, which is $20 {\times} $20 when using a cropped ResNet backbone~\cite{he2016deep}. This choice also allows for the use of larger batch sizes.

\vspace{3pt}
\noindent\textbf{Inference image size.}
Considering that many benchmarks contain images of varying sizes and aspect ratios---and often at higher resolutions---  we resize the images to a resolution slightly higher than $320$ pixels while preserving the original aspect ratio. This approach maintains the integrity of the scenes by keeping the original aspect ratio, and allows the learned queries to interact with bigger feature maps. In \cref{tab:im_size} we perform testing at different image sizes ($288, 320, 384, 432 \text{ and } 480$), using a BoQ model trained with $320{\times}320$ images. As we can see, when the images are resized to a height of 
$384$ pixels, there is a consistent improvement in Recall@1 across almost all datasets. This experiment suggests that heights of $384$ and $432$ may represent an optimal balance between image detail and the model's capacity to extract and utilize informative features. Note that the performance gains from resizing to these heights are marginal compared to the baseline size of $320$ pixels.

\setlength{\tabcolsep}{4pt}
\begin{table}[h]
\centering
\resizebox{\columnwidth}{!}{%
\begin{tabular}{@{}cccccc@{}}
\toprule
\begin{tabular}[c]{@{}c@{}}Inference \\ im. height\end{tabular} & AmsterTime & Eynsham & St-Lucia & SVOX (all) & Nordland** \\ \midrule
$288$                                                             & $48.3$       & $90.6$    & $99.8$     & $98.5$       &   $74.8$         \\
$320$                                                             & $50.0$       & $91.0$    & $99.7$     & $98.5$       &   $75.4$         \\
$384$                                                             & $53.1$       & $91.3$    & $99.9$     & $98.5$       &   $73.4$         \\
$432$                                                             & $51.4$       & $91.5$    & $99.5$     & $98.6$       &   $70.0$         \\
$480$                                                             & $50.1$       & $91.6$    & $99.4$     & $98.5$       &   $65.9$         \\ \bottomrule
\end{tabular}%
}
\caption{Recall@1 performance on various benchmarks, testing with images resized to varying heights (in pixels) while preserving their original aspect ratio. The model was trained on GSV-Cities using a fixed image size of $320 {\times} 320$.}
\label{tab:im_size}
\end{table}

\vspace{3pt}
\noindent\textbf{Data Augmentation.}
For data augmentation, we adopted the same strategy used in~\cite{ali2022gsv, ali2023mixvpr}, employing RandAugment~\cite{cubuk2020randaugment} with $N=3$, which specifies the number of random transformations to apply sequentially.

\vspace{3pt}
\noindent\textbf{Training Time.}
The training of our model, incorporating a ResNet-50 backbone~\cite{he2016deep} and two BoQ blocks (as depicted in \cref{fig:arch_2BOQ}) on GSV-Cities dataset~\cite{ali2022gsv}, takes approximately $6$ hours on a $2018$ NVidia RTX $8000$, with the power consumption capped at $180$ watts.

\section{Interpretability of the learned queries}
In this section, we demonstrate how the learned queries in BoQ can be visually interpreted through their direct interactions with the feature maps via cross-attention mechanisms. To do so, we examine their behavior in images of the same location under viewpoint changes, occlusions, and varying weather conditions. 

\begin{figure*}
  \centering
  \includegraphics[width=0.9\linewidth]{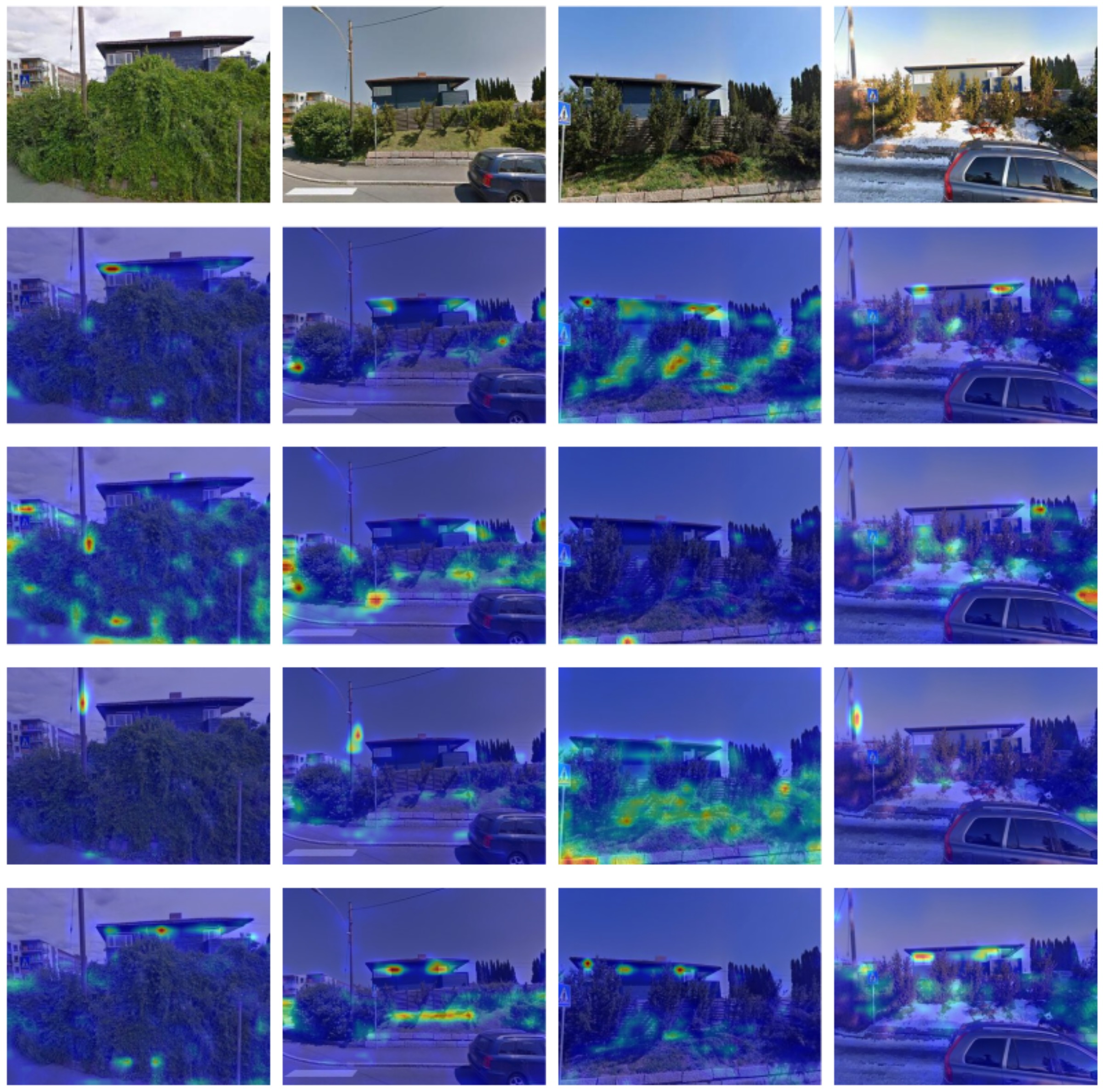}
  \caption{\textbf{Weather and occlusions.} The first row displays four images of the same location captured at different times, illustrating changes in the environment. Subsequent rows reveal the cross-attention scores between one learned query and the feature maps of the respective input image. In these heatmaps, regions with higher attention scores are indicated in warmer colors (red/yellow), signifying areas where the query is focusing more intensely. 
  First row shows four images of the same place accross different times. The following four rows show the cross-attention scores of four selected learned queries on the feature maps of the input image.}
  \label{fig:interp_1}
\end{figure*}

\begin{figure*}
  \centering
  \includegraphics[width=1\linewidth]{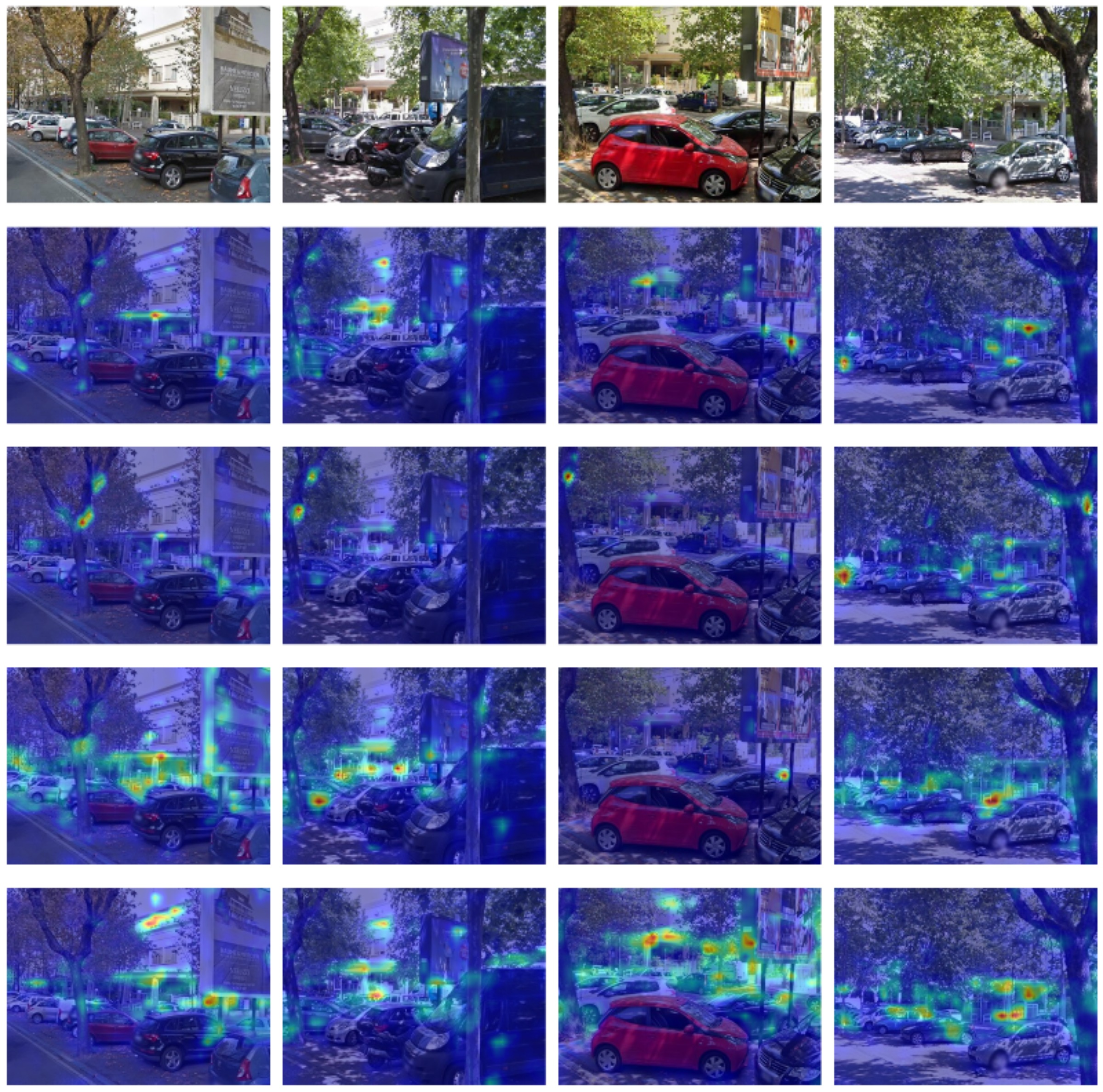}
  \caption{\textbf{Moving objects.}  The consistency of attention allocation across scenes with different moving objects (cars) underscores our model's capability in distinguishing between transient and persistent features (trees, buildings) within an urban environment.}
  \label{fig:interp_3}
\end{figure*}

\begin{figure*}
  \centering
  \includegraphics[width=1\linewidth]{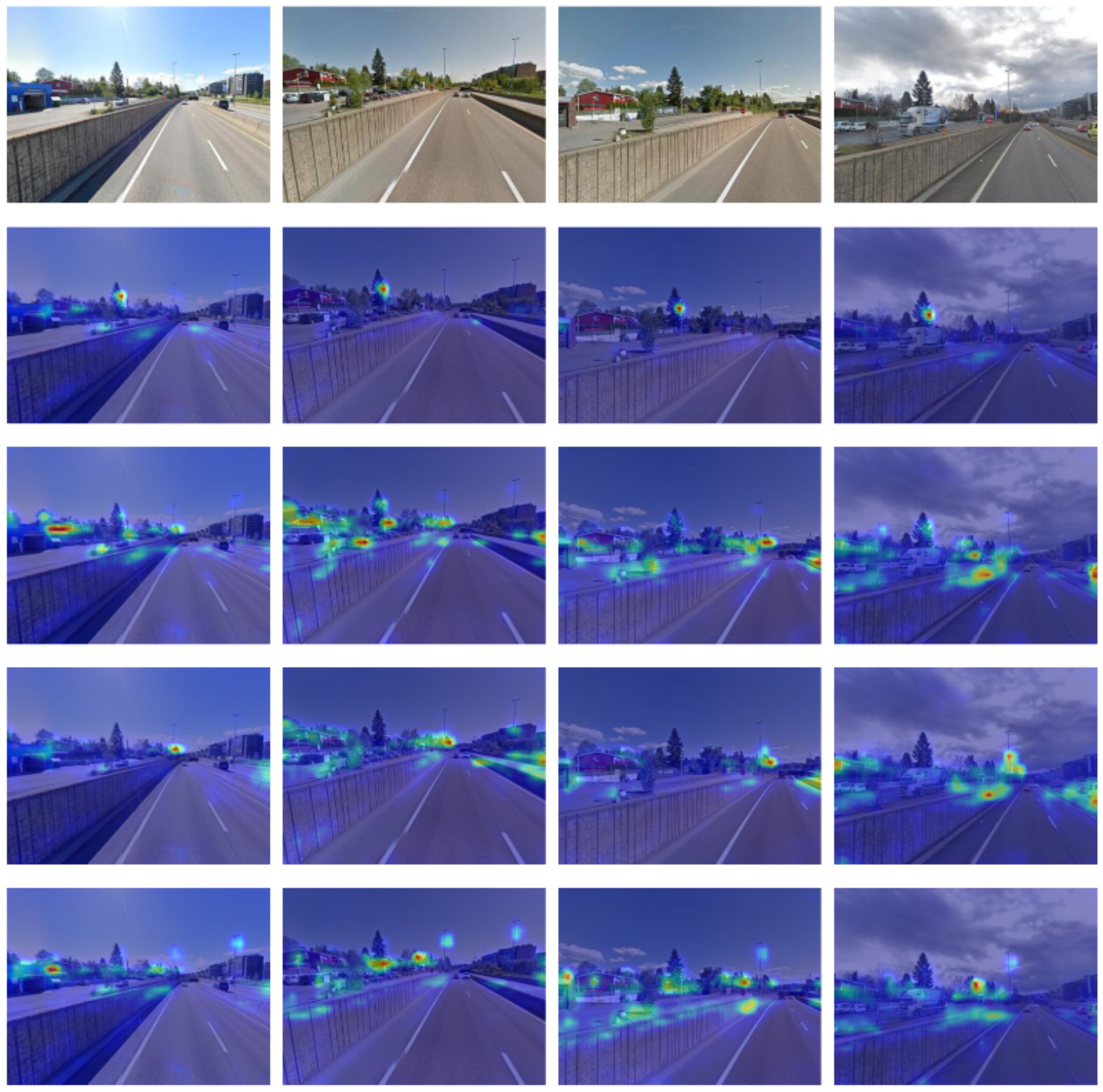}
  \caption{In this example, we can see that each query specializes in identifying particular elements within the scenes. The first query (second row) predominantly activates over big blobs of vegetation, while the second query (third row) demonstrates higher activation over architectural structures, such as buildings. These attention patterns suggest a high degree of specialization in the learned queries, enabling precise feature discrimination within complex environments.}
  \label{fig:interp_2}
\end{figure*}

The cross-attention mechanisms in our BoQ model have been instrumental in achieving fine-grained feature discrimination, as demonstrated by \cref{fig:interp_1}, \cref{fig:interp_3} and \cref{fig:interp_2}. These figures provide a visualization of the learned queries' attention patterns across diverse urban scenes and under various environmental conditions. 

\cref{fig:interp_1} demonstrates the model's temporal robustness, displaying consistent attention across images of the same location captured at different times. The learned queries reliably focus on distinctive features, such as buildings, foliage, and poles, despite variations in viewpoint, lighting, and weather conditions.

Moving objects within a scene often pose a challenge for VPR techniques. Nonetheless, as shown in \cref{fig:interp_3} our learned queries focus their attention towards static elements of the environment, avoiding moving objects like vehicles. 

\cref{fig:interp_2} underscores the specialization of the learned queries, showcasing their selective focus on relevant features, such as vegetation and buildings. This selective attention  is indicative of our model's ability to interpret complex visual information within the environment.

\end{document}